\documentclass{article}

\usepackage{arxiv}

\usepackage[utf8]{inputenc} % allow utf-8 input
\usepackage[T1]{fontenc}    % use 8-bit T1 fonts
\usepackage{hyperref}       % hyperlinks
\usepackage{url}            % simple URL typesetting
\usepackage{booktabs}       % professional-quality tables
\usepackage{amsfonts}       % blackboard math symbols
\usepackage{nicefrac}       % compact symbols for 1/2, etc.
\usepackage{microtype}      % microtypography
\usepackage{lipsum}		% Can be removed after putting your text content
\usepackage{float}
\usepackage{graphicx}
\usepackage{natbib}
\usepackage{doi}
\usepackage{siunitx}
\usepackage{multirow}

\usepackage{mathtools}

\DeclarePairedDelimiter\floor{\lfloor}{\rfloor}
\usepackage{xcolor}

\usepackage[labelformat=simple]{subcaption}

\renewcommand\thesubfigure{\alph{subfigure}}
\DeclareCaptionLabelFormat{subcaptionlabel}{\normalfont(\textbf{#2}\normalfont)}
\title{Identification of Social-Media Platform of Videos through the Use of Shared Features}

\author{ \href{https://orcid.org/0000-0001-7969-7821}{\includegraphics[scale=0.06]{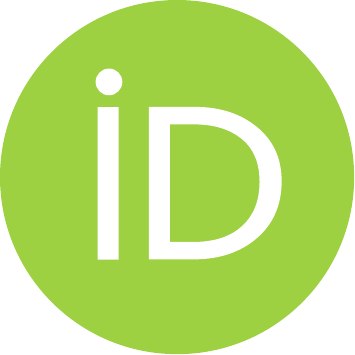}\hspace{1mm}Luca Maiano} \\
    Department of Computer, Control \\
    and Management Engineering \\
    Sapienza University of Rome\\ 
	Rome, Italy \\
	\texttt{maiano@diag.uniroma1.it} \\
	\And
	\href{https://orcid.org/0000-0002-6461-1391}{\includegraphics[scale=0.06]{orcid.pdf}\hspace{1mm}Irene Amerini} \\
	Department of Computer, Control \\
    and Management Engineering \\
    Sapienza University of Rome\\ 
	Rome, Italy \\
	\texttt{amerini@diag.uniroma1.it} \\
	\And
	\href{https://orcid.org/0000-0003-3809-8886}{\includegraphics[scale=0.06]{orcid.pdf}\hspace{1mm}Lorenzo Ricciardi Celsi} \\
	Elis Innovation Hub\\ 
	Rome, Italy\\
	\texttt{l.ricciardicelsi@ELIS.ORG} \\
	\And
	\href{https://orcid.org/0000-0001-9183-7911}{\includegraphics[scale=0.06]{orcid.pdf}\hspace{1mm}Aris Anagnostopoulos} \\
	Department of Computer, Control \\
    and Management Engineering \\
    Sapienza University of Rome\\ 
	Rome, Italy \\
	\texttt{aris@diag.uniroma1.it} \\
}

\date{}

\renewcommand{\headeright}{}
\renewcommand{\undertitle}{}
\renewcommand{\shorttitle}{}

\hypersetup{
    pdftitle={A template for the arxiv style},
    pdfsubject={q-bio.NC, q-bio.QM},
    pdfauthor={David S.~Hippocampus, Elias D.~Striatum},
    pdfkeywords={First keyword, Second keyword, More},
}

\begin{document}
\maketitle

\begin{abstract}
	Videos have become a powerful tool for spreading illegal content such as
    military propaganda, revenge porn, or bullying through social networks. To counter
    these illegal activities, it has become essential to try new methods to verify the
    origin of videos from these platforms. However, collecting datasets large enough
    to train neural networks for this task has become difficult because of the privacy
    regulations that have been enacted in recent years. To mitigate this limitation,
    in this work we propose two different solutions based on transfer learning and
    multitask learning to determine whether a video has been uploaded from
    or downloaded to a specific social
    platform through the use of shared features with images trained on the same task. By transferring
    features from the shallowest to the deepest levels of the network from the image task to
    videos, we measure the amount of information shared between these two tasks.
    Then, we introduce a model based
    on multitask learning, which learns from both tasks simultaneously. The promising experimental
    results show, in particular, the effectiveness of the multitask approach. According to our
    knowledge, this is the first work that addresses the problem of social media platform identification
    of videos through the use of shared features.
\end{abstract}

% keywords can be removed

\captionsetup[subfigure]{labelformat=simple}
\renewcommand\thesubfigure{(\alph{subfigure})}

\section{Introduction}

\label{sec:intro}
\begin{figure}[H]
\centering
{\includegraphics[width=0.7\textwidth]{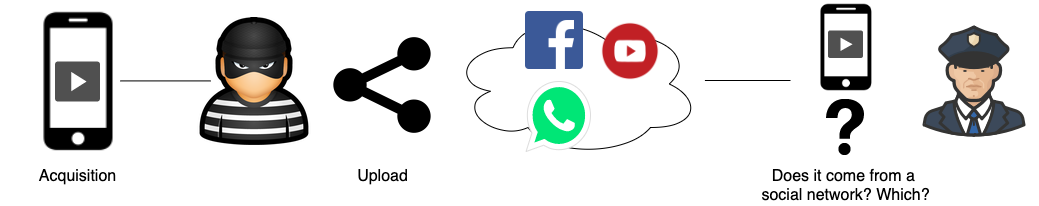}}
\caption{An application example of the proposed solution. An attacker
records a video with illegal content and shares it on social networks or
messaging apps. Subsequently, the police seize a device with this video
and want to trace the source.}
\label{fig:crime}
\end{figure}

Researchers have been studying multimedia forensics
for more than two decades in
different experimental settings; however, the practical application of
these techniques has been limited because of the high variability of real
cases, which is difficult to reproduce in experiments.
Today, the assessment of the authenticity and the source of multimedia content has become an
essential element for building trust in images and videos shared across
online platforms.
{When videos of military propaganda, revenge porn, cyberbullying, or other illegal content are shared
on social media, they can easily go viral. While it is important to immediately identify and delete
this content from social platforms, another problem to be addressed is to identify the authors of the
video to proceed with any legal action.
In many other cases, law enforcement may locate a device containing illegal content and to identify its
source, it may be necessary to understand whether the video was recorded with the hijacked device or whether
it was downloaded via messaging apps or social networks.}
% Due to the virality and ease of retrieval of these contents, being able to trace the provenance
% of a media can be very useful in contexts where images and videos are used to
% disseminate illegal content like terrorist or confidential content such as in the case of
% revenge porn or cyberbullying.
In fact, in all these cases videos and images could be used as evidence in court {and knowing how to
identify videos shared on social platforms could help identify criminal networks operating online.
}
However, for this
to be possible, it is necessary to be able to prove the origin of such content. In particular, two
problems must be solved: (1) Knowing how to reconstruct the source of acquisition (camera model or device)
and (2) understanding whether some media content found on an offending device comes from social media. Being able
to respond to the latter would allow the sharing network to be reconstructed and possible online
criminal groups to be identified. Figure~\ref{fig:crime} summarizes these two problems.

Deep learning has pushed the design of new methods that
can learn forensic fingerprints automatically from
data~\cite{Mayer_2020, Cozzolino2019_Noiseprint, huh2018fighting}, helping us to take a new step
towards applying these
techniques to real problems. Despite the promising results of
artificial neural networks,  some  limitations still remain.
Single-task learning has been very successful in computer vision
applications, with many models performing as well or even exceeding human performance
for a large number of tasks; however, they are extremely data dependent
and poorly adaptable to new contexts.
Recently, collecting data from social networks has become increasingly
difficult because of
data protection regulations and the most stringent policies introduced by the {platforms} %MDPI: 1. footnote has been put in paragraph, please confirm. 2. please add accessed date for website, please check the whole paper.
({\url{https://www.facebook.com/apps/site_scraping_tos_terms.php}, \url{https://twitter.com/en/tos}}{---accessed on 4 August  2021}).
Indeed, it is mandatory to obtain end user consent or the platform's written permission before
acquiring data via the API or web scraping of the most common social networks like Facebook, Instagram or Twitter.
Moreover, new data protection regulations, such as {GDPR} ({\url{https://europa.eu/youreurope/citizens/consumers/internet-telecoms/data-protection-online-privacy/index_en.htm}}{---accessed on 4 August  2021}),
~~~~{CCPA} \mbox{({\url{https://oag.ca.gov/privacy/ccpa}}{---accessed on 4 August  2021})}, or the Australian privacy act
are contributing to the introduction of new limitations in some countries around the world.
All these limitations make difficult to collect enough data to train a
deep-learning model.
Moreover, the human ability to
learn from experience and reuse what has been learned in new contexts
is still difficult to reproduce in machine learning as well as in multimedia forensics.
All these reasons,
along with the unavailability of large training datasets containing both
video and image content, have led
researchers to treat the problems of social-media--platform identification of
images~\cite{ameriniSNITICC2019,siddiquiSMOBITUDC2019,phanTMISSN2019,ameriniTIBSNOCBA2017}
and videos~\cite{iulianiAVFFUAMLC2019} separately.
Recently, Iuliani et al.~\cite{iulianiHRBVSI2019} showed that it is possible to
identify the source of a digital video by exploiting a reference sensor
pattern noise generated by still images taken by the same device,
suggesting that images and videos share some forensic traces.
Based on this intuition, we build a model that classifies videos from
different social-media platforms or messaging apps by taking advantage of
the shared features between images and videos.
More specifically, {to overcome the aforementioned limitations,}
we try to answer the following question:
\emph{{Is it} %MDPI: please confirm if the italic format is necessary, please check the whole paper.
possible to decide whether a video has been downloaded from a
specific social-media platform? If so, do images and videos have any
common forensic trace that can be used to solve video social-media
platform identification using both media?}
To answer these questions, we propose two methods: A method based on
transfer learning and a one based on multitask learning.
Both methods offer the possibility of reusing the features learned from one
media into another using fewer training data, a feature that is very useful in this domain
given the difficulty of finding datasets large enough to train neural networks.

In \emph{transfer learning}, we first train the base model on the image task, and
then reuse the learned features, or \emph{transfer} them, to videos. This process
tends to work if the features are general, that is, suitable to both
tasks~\cite{yosinski2014transferable}. The forensics community has adopted
widely transfer learning because,
as new manipulation methods are
continually introduced,
there is a need of
detection techniques that are
able to detect fakes with little to no training data~\cite{zhanIFBTLCNN2017, cozzolino2019forensictransfer}.
In \emph{multitask learning}, a model shares weights across multiple tasks and
makes multiple inferences in a forward pass. This method has proved to
be more scalable and robust compared to single-task models, allowing
for successful applications in several scenarios outside the forensic
community~\cite{zhang2018survey}. Some applications of multitask
learning have been even applied to multimedia forensics problems as well, for
example, to solve camera model and manipulation detection
tasks~\cite{mayerLUDFMFT2018}, as well as brand, model, and device-level
identification, using original and manipulated
images~\cite{dingCIBDKDDMTL2019}.

% Even though this is not a formal rule, the main difference between
% transfer learning and multitask learning is that the former is typically
% used when the base model has been trained on a much larger dataset with
% respect to the target task, whereas the latter is more suitable when
% applied to several tasks and datasets whose size are comparable.
We apply both learning approaches in this work to accelerate the training of a
deep-learning method for deciding whether a video has been
downloaded from a social media platform. Because the collection of
large datasets for this task is usually very difficult, if not
impossible, in practical applications because of privacy reasons, it is worth investigating
the effectiveness and the limits of transfer learning and multitasking
learning on the task of social media platform identification
of videos.

In this paper, we show how well low-level features generalize
between images and videos, demonstrating that common platform-dependent features
can be learned when the training data are not large enough to train a deep learning
model from scratch to estimate the traces left by social media platforms during the upload
phase on videos.
{The sharing process can combine multiple operations that leave different traces in the video signal.
These alterations can be exposed in various ways. For example, as first observed in~\cite{Moltisanti23234}, compression
and resizing are usually applied by Facebook to reduce the size of uploaded images and this may
happen differently on different platforms based on the resolution and size of the input data before
loading. As is widely known in multimedia forensics, such operations can be detected and characterized
by analyzing the video signal where distinctive patterns can be exhibited.
Indeed, these operations typically distort the original video signal with some artifacts that can be
detected. When the signal is used as a
source of information for the provenance analysis, different choices can
be made to preprocess the signal and
extract an effective feature representation. After the feature representation is extracted, different kinds of
machine-learning or deep-learning classifiers can then be trained to perform platform identification (see Section~\ref{sec:rationale}).}
To detect videos shared through social media platforms, we propose two methods that can learn to detect the
traces left by different social-media platforms
without any preprocessing operation on the input {frames}.
To our knowledge, this is the first work that analyzes the
similarity of the
traces left by social media platforms on images and videos used in combination.
Next, we show that the features learned in
the task of social-media identification of images can be successfully applied on
social-media identification of videos, but not vice versa,
thus suggesting a \emph{{task asymmetry}}, which could possibly be
explained by looking at social-media identification of videos as a
special case of the image task. {Indeed, as discussed in Section~\ref{sec:rationale},
shared videos may have both static and temporal artifacts, whereas shared images
have static features only.}
These findings are particularly valuable in investigative scenarios where law-enforcement agencies have to trace the origin of multimedia content without
being able to refer to other sources such as metadata. This scenario is depicted
in Figure~\ref{fig:crime}.

The rest of this paper
is organized as follows: First, in Section~\ref{sec:related-work}
we present some related work. In Section~\ref{sec:method} we describe our
methods and provide a detailed explanation of methods based on transfer learning and
multitask learning. In Section~\ref{sec:experiments} we show the
experimental results on the VISION dataset~\cite{shullaniVVIDSI2017}.
Finally, in Section~\ref{sec:conclusions} we draw the conclusions of our
work.

%%%%%%%%%%%%%%%%%%%%%%%%%%%%%%%%%%%%%%%%%%
\section{Related Work}

\label{sec:related-work}
When shared on social media platforms and messaging apps, multimedia
content is typically subjected to a series of processing and
recompression operations to speed up the loading and optimize the
display of images and videos on the platform.
Videos are typically compressed as sequence of \emph{groups of
pictures} (GOP), each of which is made by an alternation of three different kinds of frames:
\emph{I-frames}, which are not derived from any other frame and are independently encoded using a
process similar to JPEG compression, and \emph{P-frames} and
\emph{B-frames}, which are predictively
encoded using motion estimation and compensation.
While the algorithms used by social platforms are not known, all of these operations leave traces that can be
detected~\cite{ameriniTIBSNOCBA2017, ameriniSNITICC2019,
phanTMISSN2019, caldelli8553160, surveyMaiano} and, since they typically differ between different platforms~\cite{MULLAN2019S68, Pasquini2021, surveyMaiano},
they can be used to distinguish between distinct social networks.
According to the survey by Pasquini et al.~\cite{Pasquini2021}, we
can identify two main possible steps
in the digital life of a media object shared online, namely the acquisition and the
upload. Initially, a real scene is captured through an acquisition device, then,
a number of post-processing operations such as resizing, filtering, compressions, cropping,
semantic manipulations may be applied. Finally, through the upload phase, the object is
shared through social media.

Following these two steps, in the remainder of this section, we describe the state-of-the-art
methods that can be used to analyze the acquisition source or integrity of a video
(Section~\ref{sec:forensic-analysis}) and to reconstruct
information on the sharing history of a video (see \mbox{Section~\ref{sec:platform-analysis}}).

\subsection{Forensic Analysis}

\label{sec:forensic-analysis}
The main problems in traditional media forensics are the identification
of the source of images and videos and
the verification of their integrity.

Source-camera identification is the problem of tracing back the origin of a video by identifying the device or
model that captured a particular media file. This problem has been very often treated in a \emph{closed-set} setting,
meaning that all the devices that we want to be associated with a source video are known in advance.
These methods typically rely on Photo Response Non-Uniformity
(PRNU)~\cite{1634362}. Houten and Geradts~\cite{Houten200905003} propose video camera source identification of YouTube videos showing the limitations to
reach a correct identification on the shared video because of the numerous variations that affect PRNU
(e.g., compression, codec, video resolution, and changes in the aspect ratio). Similarly, another
work~\cite{Taspinar2020CameraFE} performs an analysis on stabilized and non-stabilized videos
proposing to use the spatial domain averaged frames for fingerprint extraction. A different method for PRNU
fingerprint estimation~\cite{KOUOKAM201991} takes into account the effects of video compression
on the PRNU noise selecting blocks of frames having at least one
non-null discrete cosine transform (DCT) coefficient.
Other works use PRNU to link social media profiles containing images and videos captured by the
same sensor~\cite{iulianiHRBVSI2019, Bertini2790765}.
Similar approaches have been introduced for camera model identification~\cite{rafi2019application, kuzin2018camera}.
Recently, some works have begun to deal with the problem of identifying the source of a video with \emph{limited knowledge} or
even an \emph{open-set} of devices. Cozzolino et al.~\cite{Verdoliva_2019_CVPR_Workshops} introduce a siamese method based
on~\cite{Cozzolino2019_Noiseprint} to estimate camera-based
fingerprints (called \emph{Noiseprints}) for video with no need of prior knowledge on the specific manipulation or any form of fine-tuning.
Another work~\cite{Cozzolino2020CombiningPA} from the same research group combines the PRNU and Noiseprint
to boost the performance of PRNU-based analyses based on only a few images.
In some works~\cite{iulianiAVFFUAMLC2019, yangEVIATCC2020, 8984287} video file containers have been considered for the source
identification of videos without a prior training phase. To do this, {López et al.}~%MDPI: newly added inforamtion, please confirm.
\cite{8984287} introduces a hierarchical clustering
method whereas~\cite{iulianiAVFFUAMLC2019} proposes a likelihood-ratio framework.
Mayer et al.~\cite{9054261} propose a similarity network based on~\cite{Mayer_2020} to extract features
from video patches, and to fuse multiple comparisons to produce a video-level verification decision.

Even though most of the techniques described so far are based on deep learning, which has proved successful
for camera model identification problems~\cite{Bondifstcmiwcnn}, there are other works using different
techniques. Marra et al.~\cite{Marra2016ASO} study a class of blind features based on the analysis of
the image residuals of all color bands, where no hypothesis is made on the origin of camera-specific marks,
and the identification task is regarded simply as a texture classification problem.
Chen and Stamm~\cite{Chen7368573} introduce a model of a camera's de-mosaicing algorithm by grouping together a set of submodels.
Each submodel is a nonparametric model designed to capture partial information of the de-mosaicing algorithm.
the diversity among these submodels, leads to the composition of a comprehensive representation of a
camera's de-mosaicing algorithm. Finally, an ensemble classifier is trained on the information gathered
by each sub model to identify the model of an image's source camera.

The application of forgery detection methods on shared videos has been very limited to date.
Iuliani et al.~\cite{iulianiAVFFUAMLC2019} show that the dissimilarity between a query video
and a reference file container can be estimated to detect video forgery.
Mayer and Stamm~\cite{mayer2020exposing, Mayer_2020} propose a
graph-based representation of an image, named Forensic Similarity Graph,
to detect manipulated digital images. A forgery can be detected as a separate cluster of
patches with respect to the pristine-patches cluster in the graph.
Even if the same idea has been applied~\cite{9054261} for video source identification,
the robustness of this method has not been tested on forged videos as well.

The next section presents the methods that can be used for the second phase of the pipeline,
which is the association of the platform of origin of a video.

\subsection{Platform Provenance Analysis}

\label{sec:platform-analysis}
Social-media--platform identification has been broadly explored for
images. Amerini et al.~\cite{ameriniTIBSNOCBA2017} propose a CNN
architecture that analyzes the histograms of image DCT coefficients to
reconstruct the origin of images shared across Facebook, Flickr, Google+,
Instagram, Telegram and Twitter. Another work~\cite{ameriniSNITICC2019}
introduced a CNN-based model that was used to
fuse the information extracted from the histograms of image DCT
coefficients with a noise residual extracted from the image content
through high-pass filtering. Moreover, by combining DCT features with
metadata, Phan et al.~\cite{phanTMISSN2019} showed that is possible to
track multiple sharing on social networks by extracting the traces left
by each social network within the image file. Finally,
PRNU)can be applied as suggested by Caldelli et al.~\cite{caldelli8553160} to train a
CNN to detect the social network of origin of an image.

The proposal of social-media--platform identification techniques has been
instead quite limited for videos. Amerini et al.~\cite{AMERINI20171}
introduce a preliminary work in which they evaluate different methods to
build a fingerprint to detect video shared in social networks and
also introduce a method that generates a composite fingerprint by
resorting to the use of PRNU noise. Two recent
works~\cite{iulianiAVFFUAMLC2019, yangEVIATCC2020} introduced simple
yet effective container-based methods to identify video manipulation
fingerprints and reconstruct the operating system of the source device,
proving the robustness of the method on manipulation introduced by
social media platforms.
% However, to date, we still do not have specific
% works that solve the problem of recognizing videos downloaded from
% social platforms and that distinguish the different fingerprints left by
% social networks.
Amerini et al.~\cite{maiano} propose a two-stream neural network that analyze
I-frames and P-frames in parallel. All frames are preprocessed converting them from RGB to YUV,
and the Y-channel of each frame is used as input for the network. For P-frames, the authors subtract
the Gaussian filtered version of the frame from the Y-channel to reduce the noise in these type of frames.

Nevertheless, because these preprocessing operations can change over time, it may be necessary to
periodically learn new forensic traces for smaller training datasets.
%Until now, the application of video source identification is somewhat
%limited by the difficulty of finding enough data to train a classifier.
For this reason, in
the next section, we propose two learning techniques to train models on little data,
possibly taking advantage of what is learned on similar tasks to improve
performance and speed up the learning.

%%%%%%%%%%%%%%%%%%%%%%%%%%%%%%%%%%%%%%%%%%
\section{Proposed Method}
\label{sec:method}

In this section, we propose a theoretical analysis of what could be the traces that can be left
on videos by social media and we propose a framework for platform identification.

\subsection{The Rationale}
\label{sec:rationale}

\begin{figure}[H]
\centering
{\includegraphics[width=0.7\textwidth]{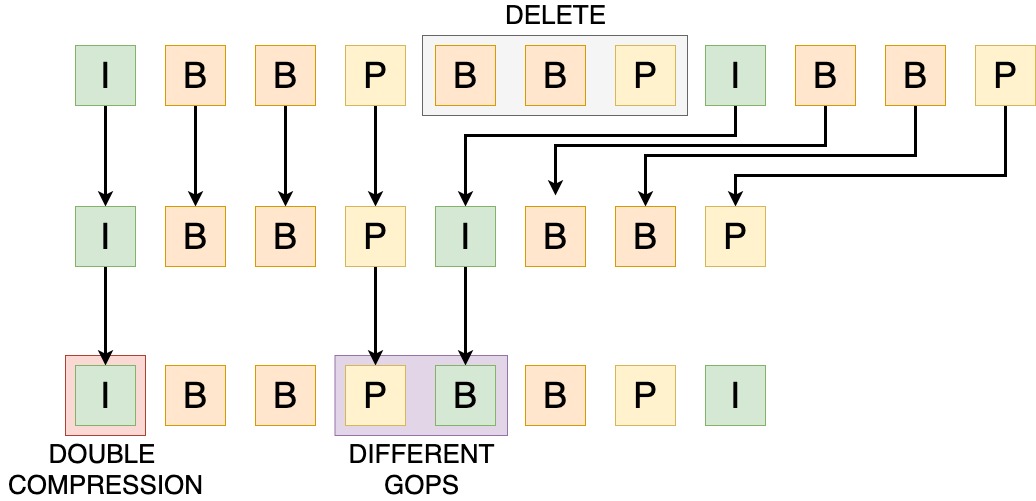}}
\caption{The top line shows an original MPEG encoded sequence. The next lines show the effect of
deleting the three frames in the shaded area. The second line shows the reordered frames and
the third line the recoded frames. The I-frame before erasing is subjected to double compression.
Some of the frames following the deletion move from one GOP sequence to another. This double
MPEG compression gives rise to specific statistical and temporal models that can be used to
identify the platform of origin.
}
\label{fig:doublecompression}
\end{figure}

{As discussed earlier, when we upload a video to a
social-media platform, it usually goes through a series of
operations, which most commonly may include recompression to reduce the bandwidth requirement for
using the video on the platform, a resize, and in some cases the removal of some frames of the video
to make it fit the maximum duration of the videos imposed by some platforms. While, as mentioned,
these operations may vary depending on the platform, in this section we want to formalize as much as
possible how these operations can leave information in the video. As shown in~\cite{popescu2005,wang1161375}, these operations can leave both static and temporal artifacts in the video signal
when a video sequence is
subjected to double MPEG compression. Statically, the I-frames of an MPEG sequence are subjected to double JPEG
compression. Temporally, frames that move from one GOP to another, as a result of frame deletion,
give rise to a relatively larger motion estimation errors. Figure~\ref{fig:doublecompression} shows an
example of a short eleven-frame MPEG sequence. In this example, during the
upload phase, the video is subjected
to the removal of three frames and subsequent recompression. The second row shows the reordered frames,
and the third line shows the re-encoded frames after recompressing the
video as an MPEG~video.}

{Statically, when an I-frame gets recompressed with
different bit rates (i.e., quantization amounts), the DCT
coefficients are subject to two quantization levels, leaving behind a specific statistical
signature in the distribution of DCT coefficients~\cite{popescu2005, mahdian1319376}. Quantization is
a pointwise operation, which can be calculated as:
\[
Q_{k}(s_1) = \floor*{\frac{k}{s_1}},
\]}
{where $s_1$ indicates the quantization step and $k$ denotes a value in the range of the input frame.
Similarly, double quantization is also a pointwise operation given by:
\[
Q_{s_1 s_2}(k) = \floor*{\floor*{\frac{k}{s_1}}\frac{s_1}{s_2}},
\]
where $s_1$ and $s_2$ are the quantization steps. From the equation above, double quantization can be
described as a sequence of three operations: A quantization with step $s_1$, a de-quantization with step $s_1$,
and a quantization with step $s_2$. As Wang and Farid show~\cite{wang1161375}, the re-quantization introduces
periodicity of the artifacts into the histograms of quantized frames. As
these artifacts will differ
depending on the quantization step used by every platform, they can be used to distinguish differences between
social media platforms.}

{Temporarily, deleting a few frames of the video to fit the maximum length set by some platforms can
in turn leave information. For example, consider deleting three frames in Figure~\ref{fig:doublecompression}.
Within the first GOP of this sequence, the I-frame and the first P-frame come from the first GOP of the original
sequence. The third B-frame, however, is the I-frame of the second GOP of the original sequence, and the second
I-frame is the first P-frame of the second GOP of the original video. When this new sequence gets re-encoded, we will
observe a larger motion error between the first and second P-frames, as they originated from different GOPs.
Furthermore, this increase in motion error will be periodic, occurring in each of the GOPs after the frame gets deleted.
Formally, consider a six-frame sequence that is encoded as $I_1, P_2, P_3, P_4, P_5, I_6$.
Because of JPEG compression and motion error, each frame can be modeled
by an additive noise, that is:
\[I_i = F_i + N_i\qquad P_j = F_j + N_j\]
with $i \neq j$, where each $N_i, N_j$ is the additional noise and $F_i, F_j$ are the
original frames. Notice that the noise for $I_1$ through $P_5$ will be
correlated to each other because they belong to the
same GOP, but not to that of $I_6$.
If we denote the motion compensation as $M(\cdot)$, we can derive the motion error for a frame $i, (i > 1)$ as:
\[\begin{split}
e_{i}&= P_i - M(I_{i-1}) \\
&= F_i + N_i - M(F_{i-1} + N_{i-1}) \\
&= (F_i - M(F_{i-1})) + (N_i - M(N_{i-1})).
\end{split}\]
Suppose now that we delete frame $P_4$, bringing frames $P_5$ and $I_6$
to the fourth and fifth positions,
respectively. $I_6$ will now be encoded as the new $P'_5$. The motion error for this new frame will be:
\[
e'_{5} = (F_6 - M(F_{5})) + (N_6 - M(N_{5})).
\]
Notice that for frames belonging to the same GOP, the components of the additive noise term $N_i - M(N_{i-1})$ are correlated,
thus, we can expect some noise cancellation. After the deletion of frame $P_4$, however, the two components of the additive
noise term $(N_6 - M(N_{5}))$ {are}  %MDPI: right bracket is newly added, please confirm.
not correlated, leading to a relatively larger motion error compared to the others.
This pattern can be learned by a deep neural network with sufficient training data samples.}

\subsection{Social Media Platform Identification Framework}

In this section, we propose two learning methods to detect and
classify different {static and temporal recompression fingerprints} left by social media platforms on shared
videos exploiting a unified set of features.
Through these learning methods, the goal is to evaluate the transferability
of features between the image and video tasks and to show the hierarchical
relation of these two tasks.
In all the following sections, we construct our methods starting from
the MISL network introduced by Bayar and
Stamm~\cite{bayarCCNNNATGPIMD2018} to train it with two different learning
approaches. This network has proven successful in several
multimedia forensics applications~\cite{Mayer_2020, mayerLUDFMFT2018},
so we decided to keep its architecture and optimize it for our setting.
Because the MISL network was originally designed to work on greyscale
images, we modified the initial constrained layer to work on RGB inputs,
therefore, we doubled the number of kernels in the first convolutional
layer from 3 to 6, to increase the expressive power of the network and
match the more complex input the model is fed with.
The network has 5 convolutional layers (called \emph{constrained},
\emph{conv1}, \emph{conv2}, \emph{conv3}, \emph{conv4}) and three
fully connected layers (called \emph{fc1}, \emph{fc2}, \emph{fc3}). The
\emph{fc3} layer has a number of neurons corresponding to the number of
output classes. The network is trained on RGB image patches for the image
social media identification platform task, and on RGB I-frame and P-frame patches
extracted from videos for the video source platform identification task.
Differently from state-of-the-art methods reported
in Section~\ref{sec:related-work}, we
decided to use the constrained convolutional layer to automatically
learn the best input transformation instead of feeding the network with
DCT histograms or reference sensor pattern noise.
Therefore, we train the network with RGB input patches extracted from video frames.

In the following sections, we use $\mathcal{I}$ and $\mathcal{V}$
to refer to the image task and video task respectively.
Moreover, we use $X_{\mathcal{I}}$ and $X_\mathcal{V}$ to refer to the input image or
video patches of the network and $Y_{\mathcal{I}}$ and $Y_{\mathcal{V}}$ to refer to the
corresponding output classes.

% \begin{figure}[H]
%     \begin{minipage}[b]{1.0\linewidth}
%         \centering
%         \centerline{\includegraphics[width=1.0\textwidth]{transfer-learning.png}}
%         \centerline{(a) Transfer learning}
%     \end{minipage}
%     \begin{minipage}[b]{1.0\linewidth}
%         \centering
%         \centerline{\includegraphics[width=1.0\textwidth]{multitask.png}}
%         \centerline{(b) Multitask learning}
%     \end{minipage}
%     \caption{Learning approaches proposed in this paper: (a) Method
%       based on transfer learning; (b) Method based on multitask
% 	learning. In the transfer-learning
%     approach we initially train the model on the image task. Then
%     we reuse the feature representations learned on images
%     to train the model on the video source platform identification task.
%     In multitask learning we share the weights of the \emph{constrained}
%     and \emph{conv1} layers of two siamese networks
%     while learning them on images and videos in parallel.}
%     \label{fig:learning}
% \end{figure}

% \subsection{Single-Task Learning}
% \label{sec:single-task}
% Single-task learning is certainly the most used approach in multimedia
% forensics and more generally in machine learning applications. In this
% section, we propose a method that classifies images and videos shared on
% social media, independently of each other.

\subsubsection{Method Based on Transfer Learning}
\label{sec:transfer-learning-method}
In this section we propose transfer learning to transfer the {static} features learned
by a base model on images to the video domain, so as to increase the performance of
the same model on this new target task.
Because we want the model to learn a certain fingerprint in
both image and video sharing tasks, we adopt this technique to measure
how features learned on one of the two tasks generalize to the other and study the
hierarchical structure of features extracted at different layers of the
network.

\begin{figure}[H]
\centering
\begin{tabular}{c}
{\includegraphics[width=.78\linewidth]{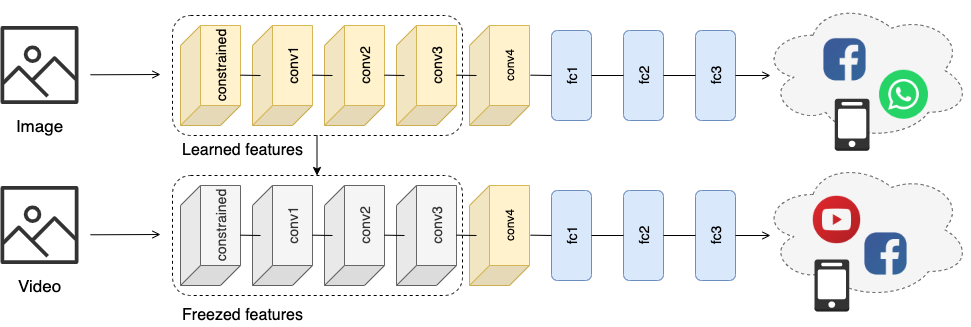}}\\
(\textbf{a}) Transfer learning \\
{\includegraphics[width=.78\linewidth]{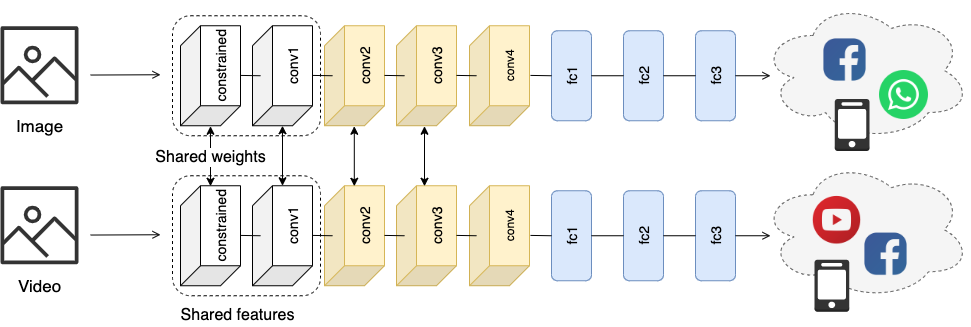}}\\
(\textbf{b}) Multitask learning
\end{tabular}

%    \begin{subfigure}[t]{1.0\linewidth}
%        \centering
%        \includegraphics[width=1.0\textwidth]{transfer-learning.png}
%        \caption{Transfer learning}\label{fig:learning-a}
%    \end{subfigure}\\ \vspace*{0.3cm}
%
%    \begin{subfigure}[t]{1.0\linewidth}
%        \centering
%        \includegraphics[width=1.0\textwidth]{multitask.png}
%        \caption{Multitask learning}\label{fig:learning-b}
%    \end{subfigure} \vspace*{0.3cm}
\caption{Learning approaches proposed in this paper: (\textbf{a}) Method
based on transfer learning; (\textbf{b}) Method based on multitask
learning. In the transfer-learning
approach we initially train the model on the image task. Then
we reuse the feature representations learned on images
to train the model on the video source platform identification task.
In multitask learning we share the weights of the \emph{constrained}
and \emph{conv1} layers of two siamese networks
while learning them on images and videos in~parallel.}
\label{fig:learning}
\end{figure}

In this setting, we initially train the model with image RGB inputs~$X_{\mathcal{I}}$ to predict
the platform of provenance $Y_{\mathcal{I}}$ of these images.
The network is initialized with a Xavier initializer~\cite{pmlr-v9-glorot10a} and
trained on $256\times256$ input patches to predict
the output classes with a cross-entropy loss function.
As shown in Figure~\ref{fig:learning}, we train this
network on native single-compressed images (i.e., images that have not been
shared on any platform) and images shared across social networks.
Next, we perform feature transfer by freezing a number of layers
from the image task and we
retrain the remaining network layers on RGB patches
$X_{\mathcal{V}}$ extracted from video frames.
We iterate this process starting
from the lower \emph{constrained} layer up to the higher \emph{fc2} layer of
the network.
At each iteration, we freeze all the middle layers in between the
constrained layer and the upper layer that we want to transfer.
\mbox{Figure~\ref{fig:learning}} shows an example of this iterative feature-transfer
approach. We initially train the model on the image task
in a single-task learning fashion to predict the
corresponding platforms of provenance. Then, we freeze all the convolutional layers
from the \emph{constrained} layer up to the \emph{conv3} layer and retrain
the remaining fully connected layers on the video task to
predict the actual new social media platforms. In \mbox{Section~\ref{sec:transfer-learning}},
we show that, according to the generic transfer learning behavior,
low-level features generalize well across the two tasks,
whereas deeper levels tend to learn more task-related representations.
This information will be useful to understand how much the two tasks share
with each~other.

\subsubsection{Method Based on Multitask Learning}
\label{sec:multitask-learning-method}
In multitask learning, we constrain some layers of two models to
learn a unique set of parameters for different tasks. In this way, we encourage
the shared layers of the network to learn a generalized representation that
should help to produce more robust and flexible classifiers {with respect to
both static and temporal features}.
As we mentioned previously, the collection large datasets of shared multimedia contents is very hard
because of several limitations (mostly related to privacy policies and API
restrictions); this approach instead helps to train
the network on smaller
training datasets. Therefore, in this setting, we force the two networks
to share a number of layers to learn more adaptable feature
representations.

Figure~\ref{fig:learning} shows the multitask learning-based network used
in this paper.
In the figure, the two proposed networks share the weights from
the \emph{constrained} layer up to the \emph{conv1} layer
to learn a common feature extractor given
input images $X_{\mathcal{I}}$ and videos $X_{\mathcal{V}}$. Next, the two networks independently
learn to predict the correct output classes $Y_{\mathcal{I}}$ and $Y_{\mathcal{V}}$. Clearly, as
suggested by the hierarchical dependencies of features maps extracted by
different layers of the network highlighted by transfer learning, for these tasks
it is not helpful to
share all the layers from the \emph{constrained} layer up to the \emph{fc2}
layer (see Section~\ref{sec:multitask-learning}). Thus, to choose
which layers to share, we use what
we have learned with transfer learning by selecting the layers that
extract the more general representations useful for both images and
videos, that is, the constrained layer and \emph{conv1} layer.

Because detecting forensics traces left by social media on videos is
harder than learning such fingerprints on images~\cite{AMERINI20171}, we train the multitask
learner by taking this information into consideration and slow down the
learning process on images. More precisely, we train the model measuring the
cross entropy loss on each task and weighing the overall loss according
to the following equation:
\begin{equation}
L = \frac{1}{N} (w_{\mathcal{I}} L_{\mathcal{I}} + w_{\mathcal{V}} L_{\mathcal{V}})
\label{eq:loss}
\end{equation}
where $L_{\mathcal{I}}$ and $L_{\mathcal{V}}$ are the cross-entropy losses on images and
videos respectively, $N$ is the number of tasks (2 in our setting), and
$w_{\mathcal{I}}$ and $w_{\mathcal{V}}$ are the weights assigned to each task. The weights
can be experimentally adjusted on each task depending on the
availability of training data and task complexity. In all our
experiments, we fix $w_{\mathcal{I}} = 0.25$ and $w_{\mathcal{V}} = 1$ such as to reduce
the loss on the image task and accelerate the improvements on videos. As
for the method based on transfer learning, at each training iteration the weights
and biases of the model are updated according to gradient descent
$w^{(\ell)} = w^{(\ell)} - \alpha \frac{\partial L_t}{\partial w^{(\ell)}}$,
where $L_t$ indicates the loss measured on task $t \in \{\mathcal{I}, \mathcal{V}\}$ and $w^{(\ell)}$
represents the weights matrix at layer~$\ell$.

\section{Experimental Evaluation}
\label{sec:experiments}
In this section, we experimentally evaluate the effectiveness of
transfer learning and multitask learning with respect to a baseline single-task
learning model fully trained on the target task. Specifically,
(1) we measure the performance of two baseline single-task models trained on
images and videos; (2) we evaluate the importance of hierarchical features with respect
to images and videos, measuring the amount of information that the two tasks share at each
level of depth through transfer learning; (3) we compare the results of
the multitask-learning approach with those relative to transfer learning
and single-\mbox{task learning.}

\subsection{Dataset and Experimental Setting}
\label{sec:dataset}
We run our experiments on the VISION dataset~\cite{shullaniVVIDSI2017}.
The dataset includes 34,427~images and 1914 videos, both in the native
format (original) and in their social media version (i.e., Facebook and WhatsApp for
images, YouTube and WhatsApp for videos), captured by 35 portable
devices of 11 major brands in many different settings. In our
experiment, we split the dataset for training and validation with a
proportion of 80\% and 10\%, respectively. Moreover, we use the
remaining 10\% of the dataset for testing purposes. All the results
reported in this section refer to this set. This ensures the robustness
of the model with respect to completely unseen data. Finally, we use
the \emph{{ffprobe}} (\url{https://ffmpeg.org/ffprobe.html}{---accessed on 4 August  2021}) analyzer to
extract the I-frames and P-frames from all videos in the dataset and
crop each frame and image into non-overlapping patches of size $H \times
W$, where $H = W = 256$.

All experiments were carried on a Google Cloud Platform n1-standard-8
instance with 8 vCPUs, 30 GB of memory, and an NVIDIA Tesla K80 GPU. The
models have been implemented using
{Pytorch} (\url{https://pytorch.org/}{---accessed on 4 August  2021}) v.1.6. For the first two sets of
experiments, we trained all the networks with the learning rate set to
\mbox{{1} $\times$ 10$^{-4}$},  %MDPI: the number has been changed to scientific notation format, please confirm.
a learning rate decay of $0.95$ fixed at every epoch, weight
decay set to {5} $\times$ 10$^{-3}$, and AdamW optimizer. In our experiments, we trained
the networks for 100 epochs with batches of size 64 and early stopping
set to 10. Finally, to train the multitask model, we set a learning rate
to {1}~$\times$~10$^{-3}$, a learning rate decay of $0.99$, and weight decay set to
{1} $\times$ 10$^{-2}$. The model was trained for 250 epochs with a batch size of 64.
All models were initialized with a Xavier initializer~\cite{pmlr-v9-glorot10a}.
%and input normalization was applied to speedup learning.

\subsection{Evaluation of Single-Task Learning}
\label{sec:e2e}
\begin{figure}[H]
\centering
\includegraphics[width=0.5\textwidth]{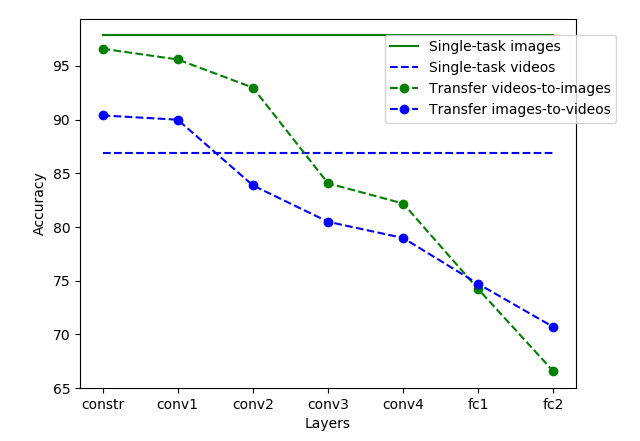}
\caption{Comparison of baseline single-task learning, transfer-learning--based, and
multitask-learning--based models
accuracy on image (in green) and video (in blue) patches.
% Single-task learning reaches
% 97.84\% and 86.85\% accuracy on images and videos respectively.
% On images, the transfer learning-based method accuracies vary from 96.60\% on the
% pretrained constrained layer to 66.56\% when all layers
% are transferred. When transfer learning is applied from the image domain to
% the video domain, the model achieves 90.39\% accuracy, gaining 3.54\% accuracy with
% respect to the single-task
% model. Multitasking learning is more sensitive to the number of shared layers and
% achieves best performance (85.91\% accuracy on images and 80.70\%
% accuracy on videos) when \emph{constrained} and \emph{conv1} layers
% are shared.
}
\label{fig:accuracies}
\end{figure}

To measure the effect of transfer learning and multitask learning, we
introduce a baseline model trained on each task. We trained the
network on images and videos, measuring the model
effectiveness on both tasks. In single task, we achieved an accuracy of
97.84\% for RGB image patches and 86.85\% for RGB video patches
extracted from frames (see Figure~\ref{fig:accuracies}). Interestingly, we did not observe substantial
differences when training the model with both I-frame and P-frame video
versus I-frame alone. However, we decided to keep both types of frames
to help generalize the model by exposing it to as different cases as
possible. Finally, to validate our choice to train the model on
RGB patches without any preprocessing on the input, we compared the
performance of our method with the Y-channel of the input after
converting RGB to YUV, and we observed a decrease in accuracy of 1.41\%
for images and 4.2\% for videos.
% For this reason, we decided to train the model on RGB patches without
% any preprocessing on the input.
\vspace{-6pt}

Tables~\ref{table:cm-images} and~\ref{table:cm-videos} report the confusion matrices of the
single-task detectors on both tasks. Even though we do not
apply any preprocessing operation to the input patches, the model
achieves state-of-the-art performance comparable to the much more
complex FusionNET~\cite{ameriniSNITICC2019}
for the image task. Indeed, the FusionNET has 99.97\%, 98.65\%, and 99.81\% patch-level
accuracy on Facebook, WhatsApp, and native images, respectively, with an average
difference of +1.89\% with respect respect to our single-task model.
For videos, our method suffers a drop in accuracy compared to
the image task, but it still achieves results around 86.85\%. Finally,
we tested
the overall accuracy of the model at image level and video level applying \emph{majority voting}
(i.e., the class that is voted by the majority of input patches is selected as the predicted
class of the entire image or video),
reaching 98.52\% and 85.48\%, respectively.

\begin{table}[H]
\centering
\caption{Confusion matrix of the baseline single-task model on patches extracted from images.
FBH and FBL represent high-quality and low-quality patches from Facebook. WA and
NAT represent WhatsApp and native image patches respectively.}
    \begin{tabular}{l|c|c|c|c|}
        \cline{3-5}
        \multicolumn{2}{c|}{}&FB&WA&NAT\\
        \cline{2-5}
        \multirow{1}{*}{}& FB & \textbf{98.78\%} & $0.05\%$ & $1.17\%$\\
        \cline{2-5}
        \multirow{1}{*}{}& WA & $0.23\%$ & \textbf{98.37\%} & $1.40\%$\\
        \cline{2-5}
        \multirow{1}{*}{}& NAT & $1.56\%$ & $1.31\%$ & \textbf{97.13\%} \\
        \cline{2-5}
    \end{tabular}
\label{table:cm-images}
\end{table}
\vspace{-10pt}

\begin{table}[H]
\centering
\caption{Confusion matrix of the baseline single-task model on patches extracted from video
frames. YT, WA and NAT represent YouTube, WhatsApp and native video patches respectively.}
    \begin{tabular}{l|c|c|c|c|}
        \cline{3-5}
        \multicolumn{2}{c|}{}&YT&WA&NAT\\
        \cline{2-5}
        \multirow{1}{*}{}& YT & \textbf{85.28\%} & $8.36\%$ & $6.45\%$\\
        \cline{2-5}
        \multirow{1}{*}{}& WA & 11.56\% & \textbf{72.35\%} & $16.09\%$\\
        \cline{2-5}
        \multirow{1}{*}{}& NAT & 2.85\% & $11.15\%$ & \textbf{86.00\%}\\
        \cline{2-5}
    \end{tabular}
\label{table:cm-videos}
\end{table}

\subsection{Evaluation of Transfer Learning}
\label{sec:transfer-learning}

We performed a set of experiments to measure the robustness of methods based on
transfer learning to
images and videos. To perform the experiments, we froze some layers of
the network with the learned parameters in one task and we retrained the
remaining layers in the other task. To track the hierarchical
dependencies of each task and measure the similarity of the two, we
repeated this process for each level in the network from the
\emph{constrained} layer up to the \emph{fc2} layer. As shown in Figure~\ref{fig:accuracies},
the two tasks share low-level features, whereas deeper representations
are mostly related to the target task with the accuracy varying from
66.56\% to 96.60\% for images and from 70.69\% to 90.39\% for videos at
the patch level.
On images (in green), the accuracy deteriorates as more layers are shared from
the pretrained \emph{constrained} layer up to the \emph{fc2} layer.
% For videos (in blue), transfer learning begins with an accuracy of 90.39\% on the
% pretrained \emph{constrained} layer and drops to 70.69\% when all layers
% are transferred.
When knowledge is transferred from the image
domain to the video domain (in blue), the network achieves 90.39\% accuracy,
gaining 3.54\% accuracy with respect to the single-task model.
This result confirms the intuition that
lower-level features are shared between the two tasks, and that the
\emph{hierarchical} dependence between the two tasks can be used
to train a deep-learning model on a small set of images or videos
originating from social networks.
In fact, the features
extracted from the deeper levels turn out to be specific to the task
being solved and therefore less generalizable, whereas the features
extracted from the first levels of the network are more generic and,
therefore, can be shared between the two tasks.
The accuracy increases when measuring the performance at the image and at
the video level. Specifically, the accuracy on images varies from
80.15\% to 97.87\%,
with maximum accuracy up to 98.37\% obtained by transferring video features up
to the \emph{conv2} layer.
Finally, when transferring
from images to video, we can observe an increase in accuracy from 85.48\%
to 92.61\% on the video
classifier, but the same does not happen for the transfer from video to
images. This result can probably be explained by considering the videos
as a more specific case and then thinking of this task as a subset of
the corresponding task on images, thus suggesting an \emph{asymmetry}
between the two tasks.

\subsection{Evaluation of Multitask Learning}
\label{sec:multitask-learning}

\begin{figure}[H]
\centering
\includegraphics[width=0.4\textwidth]{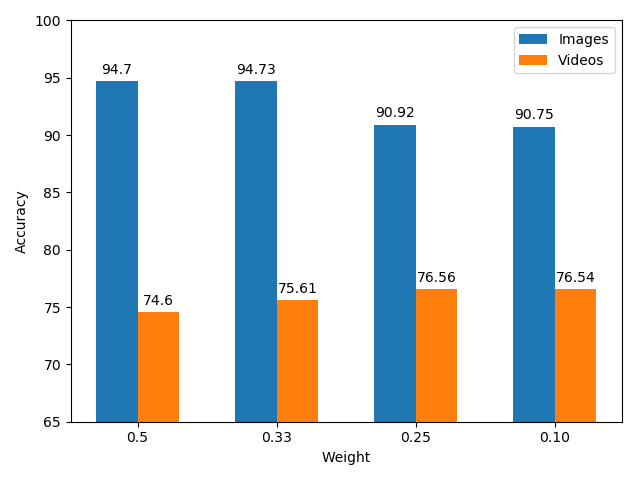}
\caption{Test accuracy of the multitask learner on images and video
I-frames obtained by fixing $w_{\mathcal{V}} = 1$ and varying the images weight
$w_{\mathcal{I}}$ according to Equation~\eqref{eq:loss}.}
\label{fig:weights}
\end{figure}

With this last experiment, we measured the performance of the proposed multitask
learner. Specifically, we chose to train two networks on both
tasks, by forcing them to share weights between the first two
convolutional layers, namely the \emph{constrained} and \emph{conv1}
layers. Because of the different complexity of the two tasks highlighted by
transfer learning, it is not useful to share all the layers between the
two networks and it becomes necessary to balance the learning speed on
images with compared to the videos.
Therefore, we initially run several experiments
with variable weighted loss according to Equation~\eqref{eq:loss}.
To speed up the training, in this exploratory phase we chose to train the networks
on images and I-frames only for the videos. We report the results of this experiment
in Figure~\ref{fig:weights}. We have varied the images weight $w_{\mathcal{I}}$ from $0.5$ down to~$0.1$.
Then, we chose $w_{\mathcal{I}} = 0.25$ so as to
maximize the accuracy of the multitask learner on the video task and we
retrained the multitask-learning-based model sharing the \emph{constrained} and \emph{conv1} layers between
the two tasks.
In this configuration, the multitask-learning-based model achieved 85.91\% accuracy on
images and 81.70\% accuracy on videos.
Finally, we tested the overall accuracy of the model
at the image and the video level, reaching 92.08\% and 91.55\% accuracy
on the images and the videos
respectively. In this setting the model reaches an
accuracy comparable to the single-task learner for the video task.

\vspace{-6pt}

{To evaluate the performance of our method, we compared it with the state-of-the-art two-stream network
introduced by Amerini et al.~\cite{maiano}. To compare the performance of the
transfer-learning and multitask-learning--based methods with that of
Amerini et al.~\cite{maiano}, we retrained the model of that work in this new
setting. {Table}~\ref{table:comparison} %MDPI: tables should be cited in numerical order, table 3 should be cited before table 4, please revise.
shows the results of this
comparison. Splitting the dataset at video level instead of frame level, the
method from Amerini et al.~\cite{maiano} records a drop in
accuracy of 15.47\% compared to the configuration used in the original
paper.}

\begin{table}[H]
\centering
\caption{Comparison of video patch classification accuracy of our
transfer-learning and multitask-learning methods with the one of
Amerini et al.~\cite{maiano} on the VISION dataset.}
\label{table:comparison}
\begin{center}
\begin{tabular}{ |c|c| } 
 \hline
\textbf{Method} & \textbf{Accuracy} \\ 
\hline
~\cite{maiano} & 80.04\% \\

TL (ours) & \textbf{{90.39}\%}\\

MT (ours) & 81.70\%\\
 \hline
\end{tabular}
\end{center}
\end{table}

\section{Discussion}
\label{sec:discussion}

While the method based on transfer learning achieves a higher overall accuracy than the
one based on multitask learning, we investigated the different performance of
these two approaches. To analyze and compare the results of the two
methods, we kept the
best configuration of the multitask learning-based model and examined the
results of the transfer learning-based model when transferring features from the \emph{constrained}
and \emph{conv1} layers as for the multitask network.
Table~\ref{tab3}
shows the confusion matrices of these two methods on videos.

First, the transfer-learning model is able to achieve better results
than the baseline model on YouTube and native videos (see Tables~\ref{table:cm-videos}
and~\ref{tab3}a). However, the WhatsApp class gets more easily
confused with the other classes. Second, the multitask learner
(\mbox{Table~\ref{tab3}b}) tends to learn
features representations that are more equally separated, with accuracy on all classes
that oscillates between 79.25\% and 83.68\%, making the multitask learner less biased
and more robust across all the classes. Moreover, thanks to this property,
the multitask approach introduces an improvement in
classification performance on WhatsApp compared to transfer learning (+{10.74}\%,  %MDPI: comma has been changed to decimal dot, please confirm.
see
Table~\ref{tab3}) and
the baseline model (+{7.89}\%, see Tables~\ref{table:cm-videos}
and~\ref{tab3}b). Because WhatsApp is the only class shared by the image
and video sets, it might suggest that training a model in a multitask setting on
images and videos from the same social media platform could be even more beneficial.
To evaluate this intuition we tested the model on WhatsApp with native
images and videos, achieving
encouraging results. The multitask-learning model achieves higher
accuracy than transfer learning and single-task learning, again
obtaining more stable accuracy
across all classes. {Most likely, images and videos shared through the same platform use very similar
compression algorithms, favoring the learning of the alterations introduced when the content is
recompressed when uploaded to the platform.}
Table~\ref{tab5}b,c show the results of this experiment.
However, because of the lack of publicly available datasets containing both images and
videos we are not able to verify whether this is the case with more classes and leave this issue
open for future research.

\begin{table}[H]
\caption{{Confusion} matrices on video patches of the transfer-learning (a) and multitask
learning (b) models sharing the \emph{constrained} and \emph{conv1} layers.\label{tab3}} 
\begin{minipage}[b]{1.0\linewidth}
        \centering
        \begin{tabular}{l|c|c|c|c|}
            \cline{3-5}
            \multicolumn{2}{c|}{}&FB&WA&NAT\\
            \cline{2-5}
            \multirow{1}{*}{}& YT & \textbf{91.24\%} & $1.08\%$ & $7.66\%$\\
            \cline{2-5}
            \multirow{1}{*}{}& WA & 13.33\% & \textbf{69.50\%} & $17.15\%$\\
            \cline{2-5}
            \multirow{1}{*}{}& NAT & 6.05\% & $1.49\%$ & \textbf{92.45\%}\\
            \cline{2-5}
        \end{tabular}
        \centerline{(a) Transfer learning.}
        \label{table:tl-images}
    \end{minipage}\\

   \begin{minipage}[b]{1.0\linewidth}
        \centering
        \begin{tabular}{l|c|c|c|c|}
            \cline{3-5}
            \multicolumn{2}{c|}{}&FB&WA&NAT\\
            \cline{2-5}
            \multirow{1}{*}{}& YT & \textbf{83.68\%} & $6.19\%$ & $10.04\%$\\
            \cline{2-5}
            \multirow{1}{*}{}& WA & 10.04\% & \textbf{80.24\%} & $9.72\%$\\
            \cline{2-5}
            \multirow{1}{*}{}& NAT & 10.58\% & $10.17\%$ & \textbf{79.25\%}\\
            \cline{2-5}
        \end{tabular}
        \centerline{(b) Multitask learning.}
        \label{table:mt-images}
    \end{minipage} 
\end{table}
\unskip
\begin{table}[H]
\caption{Confusion matrices on video patches of the single-task learning (a),  transfer-learning (b) and multitask
learning (c) models sharing the \emph{constrained} and \emph{conv1} layers.\label{tab5}}
\begin{minipage}[b]{1.0\linewidth}
        \centering
        \begin{tabular}{l|c|c|c|}
            \cline{3-4}
            \multicolumn{2}{c|}{}&WA&NAT\\
            \cline{2-4}
            \multirow{1}{*}{}& WA  & \textbf{60.12\%} & $39.88\%$\\
            \cline{2-4}
            \multirow{1}{*}{}& NAT & $28.07\%$ & \textbf{71.93\%} \\
            \cline{2-4}
        \end{tabular}
        \centerline{(a) Single-task learning.}
        \label{table:st-videos}
    \end{minipage}\\
\begin{minipage}[b]{1.0\linewidth}
        \centering
        \begin{tabular}{l|c|c|c|}
            \cline{3-4}
            \multicolumn{2}{c|}{}&WA&NAT\\
            \cline{2-4}
            \multirow{1}{*}{}& WA  & \textbf{63.08\%} & $36.92\%$\\
            \cline{2-4}
            \multirow{1}{*}{}& NAT & $23.69\%$ & \textbf{76.30\%} \\
            \cline{2-4}
        \end{tabular}
        \centerline{(b) Transfer learning.}
        \label{table:tl-videos}
    \end{minipage}\\

    \begin{minipage}[b]{1.0\linewidth}
        \centering
        \begin{tabular}{l|c|c|c|}
            \cline{3-4}
            \multicolumn{2}{c|}{}&WA&NAT\\
            \cline{2-4}
            \multirow{1}{*}{}& WA & \textbf{71.48\%} & $28.52\%$\\
            \cline{2-4}
            \multirow{1}{*}{}& NAT & $26.16\%$ & \textbf{73.84\%} \\
            \cline{2-4}
        \end{tabular}
        \centerline{(c) Multitask learning.}
        \label{table:mt-videos}
    \end{minipage} 
\end{table}

\section{Conclusions}
\label{sec:conclusions}
In this paper, we propose two methods to identify the platform of origin of videos shared
on different social networks through the
use of joint features from images.
Moreover, we show that images and videos share common forensic traces and a mixed
approach may be beneficial in some cases where data are not enough to
train a single-task model. By applying a transfer-learning--based method
on both tasks, we experimentally showed that: (1) As
expected, low-level features generalize well
across images and videos, whereas deeper-feature mappings are more related to the target task,
therefore suggesting that a common feature hierarchy exists between the two tasks;
(2) image features can be successfully used to identify the social media
platform in which videos have been uploaded, helping to improve
performance over single task learning.
Finally, we showed the promising effectiveness of a multitask-learning
approach compared to
single-task learning. In this way, the model can learn from images
and videos simultaneously,
learning more generic and robust features across all classes.
These findings
suggest that learning from multiple media could help to
overcome the hurdle of training low-data models, by taking advantage of the similarity of
different forensic tasks, usually treated separately.

Future work could be aimed at gathering a larger training dataset for
social-media--platform identification of multimedia content and at studying the case of multiple sharing
considering both images and videos. Moreover, a limitation of our method is that it
appears susceptible to false positive classifications, leaving room for
improvement.

%%%%%%%%%%%%%%%%%%%%%%%%%%%%%%%%%%%%%%%%%%
\vspace{6pt}

%%%%%%%%%%%%%%%%%%%%%%%%%%%%%%%%%%%%%%%%%%
%% optional
%\supplementary{The following are available online at \linksupplementary{s1}, Figure S1: title, Table S1: title, Video S1: title.}


\section*{Acknowledgements}
\label{sec:acknowlegment}
%%%%%%%%%%%%%%%%%%%%%%%%%%%%%%%%%%%%%%%%%%
Partially supported by Supported by the ERC Advanced Grant 788893 AMDROMA and the EC H2020 RIA project "SoBigData++" (871042).
%%%%%%%%%%%%%%%%%%%%%%%%%%%%%%%%%%%%%%%%%%



\end{document}